\documentclass[runningheads]{llncs}

\usepackage{amsmath,amssymb}
\usepackage[T1]{fontenc}
\usepackage{hyperref}
\usepackage{threeparttable}
\usepackage{booktabs}
\usepackage{multirow}
\usepackage{amsmath}
\usepackage{algorithm}
\usepackage{algpseudocode}
\usepackage{graphicx}
\begin{document}
\title{Adversarial Instance Generation and Robust Training for Neural Combinatorial Optimization with Multiple Objectives}

\author{
Wei Liu\inst{1} \and
Yaoxin Wu\inst{2} \and
Yingqian Zhang\inst{2} \and
Thomas B{\"a}ck\inst{1} \and
Yingjie Fan\inst{1}
}

\authorrunning{W. Liu et al.}

\institute{
LIACS, Leiden University, Leiden, The Netherlands\\
\email{\{w.liu,t.h.w.baeck,y.fan\}@liacs.leidenuniv.nl}
\and
Eindhoven University of Technology, Eindhoven, The Netherlands\\
\email{\{y.wu2,yqzhang\}@tue.nl}
}
\maketitle              
\begin{abstract}
Deep reinforcement learning (DRL) has shown great promise in addressing multi-objective combinatorial optimization problems (MOCOPs). Nevertheless, the robustness of these learning-based solvers has remained insufficiently explored, especially across diverse and complex problem distributions. In this paper, we propose a unified robustness-oriented framework for preference-conditioned DRL solvers for MOCOPs. Within this framework, we develop a preference-based adversarial attack to generate hard instances that expose solver weaknesses, and quantify the attack impact by the resulting degradation on Pareto-front quality. We further introduce a defense strategy that integrates hardness-aware preference selection into adversarial training to reduce overfitting to restricted preference regions and improve out-of-distribution performance.
The experimental results on multi-objective traveling salesman problem (MOTSP), multi-objective capacitated vehicle routing problem (MOCVRP), and multi-objective knapsack problem (MOKP) verify that our attack method successfully learns hard instances for different solvers. Furthermore, our defense method significantly strengthens the robustness and generalizability of neural solvers, delivering superior performance on hard or out-of-distribution instances.

\end{abstract}

\section{Introduction}
Multi-objective combinatorial optimization problems (MOCOPs) have been extensively studied in computer science and operations research~\cite{Liu2020MOMetaheuristics}, and arise in a wide range of real-world applications, including transportation scheduling and scheduling~\cite{ZajacHuber2021MORouting}, feature selection~\cite{SainiSaha2021Survey}, energy planning~\cite{Cui2017EnergySaving}, and communication network optimization~\cite{Fei2017MOWSN}. 
Rather than seeking a single optimal solution as in single-objective optimization, MOCOPs involve multiple conflicting objectives and therefore aim to identify a set of Pareto-optimal solutions, known as the Pareto set. A good approximation to the Pareto set should achieve both convergence and diversity. Due to the inherent NP-hardness of combinatorial optimization and the presence of multiple objectives, most MOCOPs are difficult to solve in practice.  

Although classic methods, including exact \cite{ehrgott2016exact} and heuristic algorithms \cite{herzel2021approximation}, have long achieved competitive performance in obtaining an approximate Pareto set, they typically require domain-specific knowledge or massive iterative search. Instead, deep reinforcement learning (DRL) has emerged as a transformative approach to address MOCOP. Unlike supervised learning \cite{zoph2017neural}\cite{gasse2019exact}, DRL eliminates the dependence on labeled datasets. Moreover, compared to heuristic and exact algorithms, DRL demonstrates superior efficiency in identifying near-optimal solutions in a reasonable computational time~\cite{NEURIPS2021_3d863b36}. 

However, despite these advantages, the robustness of DRL solvers for MOCOPs remains considerably underexplored.
For single-objective combinatorial optimization problems, prior studies already investigated robustness by conducting experiments on instances drawn from non-i.i.d. distributions and proposed mitigation strategies such as modified training procedures and refined model architectures \cite{zhang2022learning,lu2023roco}.
When training and test distributions diverge, DRL models may exhibit substantial performance drops on out-of-distribution instances in MOCOP scenarios, potentially due to overfitting to spurious, distribution-specific cues, a phenomenon often discussed as shortcut learning in deep neural networks~\cite{geirhos2020shortcut}.
This motivates our work on enhancing the out-of-distribution generalizability of DRL solvers for MOCOPs.

In this paper, we propose a robustness-oriented framework for this setting. Unlike existing neural approaches that primarily target in-distribution performance, we systematically evaluate how distribution shifts and preference conditioning affect the performance of DRL solvers. In addition, we introduce adversarial instance generation and robust training strategies to improve generalization across diverse instance distributions. Our main contributions are as follows:
\begin{itemize}
    \item  
   We introduce a Preference-based Adversarial Attack (PAA) method to target DRL models for MOCOPs.
    PAA undermines DRL models by generating hard instances that degrade solutions of subproblems associated with specific preferences. The generated instances effectively lower the quality of the Pareto fronts in terms of hypervolume.

    \item 
    We propose a Dynamic Preference-augmented Defense (DPD) method to mitigate the impact of adversarial attacks. By integrating a hardness-aware preference selection strategy into adversarial training, DPD effectively alleviates the overfitting to restricted preference spaces. It enhances the robustness of DRL models, thereby promoting their generalizability across diverse distributions.

    \item 
    We evaluate our methods on three classical MOCOPs: multi-objective traveling salesman problem (MOTSP), multi-objective capacitated vehicle routing problem (MOCVRP), and multi-objective knapsack problem (MOKP). The PAA method substantially impairs competitive DRL models, while the DPD method enhances their robustness, resulting in strong out-of-distribution generalizability.
    
\end{itemize}

\section{Related Work}
\subsection{MOCOP Solvers}
MOCOP solvers are typically classified into exact, heuristic, and learning-based methods. Exact algorithms provide Pareto-optimal solutions, but become computationally intractable for large-scale problems~\cite{FLORIOS20141}~\cite{halffmann2022exact}. Heuristic methods, particularly evolutionary algorithms~\cite{zhang2007moea}~\cite{ke2014simple}~\cite{fang2020mining}, effectively explore the solution space through crossover and mutation operations~\cite{tian2021evolutionary}, generating a finite set of approximate Pareto solutions in acceptable time. However, their reliance on problem-specific, hand-crafted designs limits their applicability~\cite{4358754}. 

Learning-based solvers, particularly those based on deep reinforcement learning, have seen growing adoption in MOCOPs, largely due to their high performance and efficiency. 
Current research on DRL solvers belongs mainly to two paradigms: one-to-one and many-to-one. In the one-to-one paradigm, each subproblem is addressed by an individual neural solver~\cite{wu2020modrld}~\cite{9040280}. In contrast, the many-to-one paradigm streamlines the computational process by using a shared neural solver to handle multiple subproblems, which outperforms the one-to-one paradigm. 
Among them, the efficient meta neural heuristic (EMNH)~\cite{chen2024efficient} learns a meta-model that is rapidly adapted to each preference to solve its subproblem.
The  preference-conditioned multi-objective combinatorial optimization (PMOCO)~\cite{lin2022pareto} uses a hypernetwork to generate decoder parameters tailored to each subproblem. The conditional neural heuristic (CNH)~\cite{fan2024conditional} leverages dual attention, while the weight embedding model with conditional attention (WE-CA)~\cite{chen2025rethinking} employs feature-wise affine transformations, to enhance preference–instance interaction within the encoder. Our study demonstrates that the proposed attack and defense framework is sufficiently general to challenge and robustify models from all three categories.

\subsection{Robustness of DRL Models for Combinatorial Optimization Problems (COPs)}
Robustness COPs have been studied from both theoretical and neural perspectives. 
From the theoretical side, Varma et al. \cite{varma2021average} introduced the notion of average sensitivity, measuring the stability of algorithmic outputs under edge deletions in classical COPs such as minimum cut and maximum matching.
On the neural side,
several studies have investigated hard instance generation and defense methods to improve the robustness of DRL solvers for COPs. For example, Geisler et al.~\cite{geisler2021generalization} proposed an efficient and sound perturbation model that adversarially inserts nodes into TSP instances to maximize the deviation between the predicted route and the optimal solution. Zhang et al.~\cite{zhang2022learning} developed hardness-adaptive curriculum learning methods (HAC) to assess the hardness of given instances and then generate hard instances during training based on the relative difficulty of the solver. 
Lu et al.~\cite{lu2023roco} introduced a no-worse optimal cost guarantee (i.e., by lowering the cost of a partial problem) and generated adversarial instances through edge modifications in the graph.
In contrast to these approaches that focused on generating hard instances, Zhou et al.~\cite{zhou2024collaboration} focused on defending neural COP solvers, by an ensemble-based collaborative neural framework designed to improve performance simultaneously in both clean and hard instances.

\section{Preliminaries}
\subsection{Problem Statement}
The mathematical formulation of an MOCOP is generally given as:
\begin{equation}
\min_{\pi \in \mathcal{X}} F(\pi) = \big(f_1(\pi), f_2(\pi), \dots, f_m(\pi)\big), \tag{1}
\end{equation}
where $\mathcal{X}$ represents the set of all feasible solutions, and $F(\pi)$ is an $m$-dimensional vector of objective values. 
A solution $\pi$ to an MOCOP is considered feasible if and only if it satisfies all the constraints specified in the problem. For example, the MOTSP is defined on a graph $G$ with a set of nodes $V = \{0, 1, 2, \dots, n\}$. Each solution $\pi = (\pi_1, \pi_2, \dots, \pi_T)$ is a tour consisting of a sequence of nodes of length $T$, where $\pi_j \in V$.

For multi-objective optimization problems, the goal is to find pareto-optimal solutions that simultaneously optimize all objectives. These pareto-optimal solutions aim to balance trade-offs under different preferences for the objectives. In this paper, we use the following the Pareto concepts \cite{Qian2013}:

\noindent \textit{\textbf{Definition 1 (Pareto Dominance).}} Let $u, v \in \mathcal{X}$. The solution $u$ is defined as dominating solution $v$ (denoted $u \prec v$) if and only if, for every objective $i$ where $i \in \{1, \dots, m\}$, the objective value $f_i(u)$ is less than or equal to $f_i(v)$, and there exists at least one objective $j$ where $j \in \{1, \dots, m\}$, such that $f_j(u) < f_j(v)$.

\noindent \textit{\textbf{Definition 2 (Pareto Optimality).}} A solution $x^* \in \mathcal{X}$ is Pareto optimal if it is not dominated by any other solution in $\mathcal{X}$. Formally, there exists no solution $x' \in \mathcal{X}$ such that $x' \prec x^*$. The set of all Pareto-optimal solutions is referred to as the Pareto set $\mathcal{P} = \{x^* \in \mathcal{X} \mid \nexists x' \in \mathcal{X} \text{ such that } x' \prec x^*\}$. The projections of Pareto set into the objective space constitute Pareto front $\mathcal{PF} = \{F(x) \mid x \in \mathcal{P}\}$.




\subsection{Glimpse of Robustness of DRL Solvers}
\label{subsec:robustness-drl}
\begin{figure}[t]
    \centering
    \includegraphics[width=0.7\textwidth]{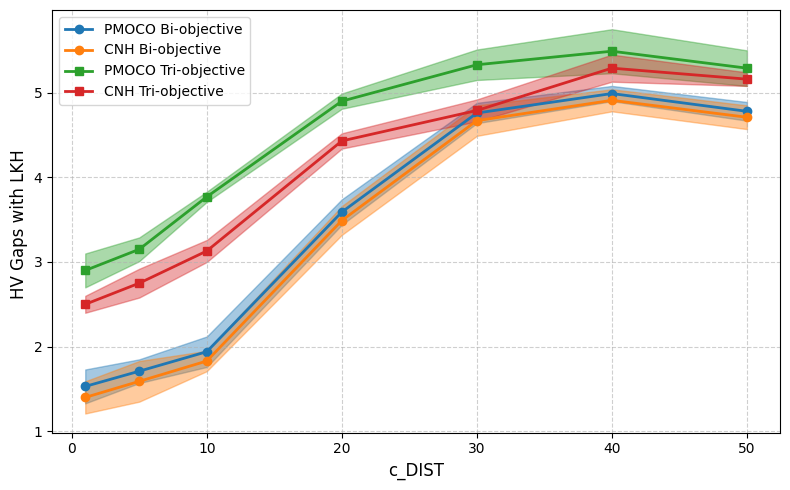}
    \caption{Results for Varying $c_{\text{DIST}}$ in Gaussian Mixture Generator.}
    \label{preliminary}
\end{figure}
Given the unexplored robustness of DRL models for MOCOPs, we first examined the performance of two representative neural solvers, PMOCO~\cite{lin2022pareto} and CNH~\cite{fan2024conditional}, for MOTSP. Both solvers are pretrained on clean 50-node bi-/tri-objective TSP instances following their original training protocols.

Out-of-distribution test instances are generated using a Gaussian mixture model (GMM) generator. We vary $c_{\text{DIST}} \in \{1, 5, 10, 20, 30, 40, 50\}$, where $c_{\text{DIST}}$ controls the spatial spread of the clusters, which determines the hardness of the instances. WS-LKH \cite{tinos2018efficient} is used as a baseline for comparison. Figure~\ref{preliminary} illustrates the HV gaps, representing the difference between the solutions produced by a neural solver and those using WS-LKH: 
\begin{equation}
\text{Gap} = \frac{HV_{\text{LKH}} - HV_{\text{DRL}}}{HV_{\text{LKH}}} \times 100.
\end{equation}

 The results reveal that with increasing $c_{\text{DIST}}$, 
 the performance of the neural solvers deteriorates and the gap between their solutions and those provided by WS-LKH widens.
This trend highlights a significant limitation that neural solvers trained on uniformly distributed instances struggle to maintain robustness and get high-quality solutions in response to where test instances become more diverse and complex.

\section{The Method}
\begin{figure*}[t] 
    \vspace{-10pt} 
    \centering
    \includegraphics[width=1\textwidth]{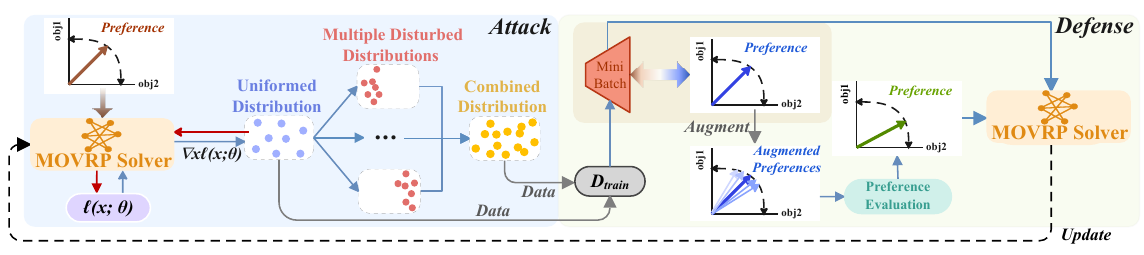}
    \vspace{-10pt} 
    \caption{Attack and Defense of Neural Solvers for MOCOP.}
    \label{fig:pic2}
\end{figure*}

In this section, we introduce a preference-based adversarial attack method that generates instances to degrade solver performance and assess robustness. Furthermore, we propose a dynamic preference-augmented defense method to robustify neural solvers. The general framework is illustrated in Figure~\ref{fig:pic2}.
\subsection{Preference-based Adversarial Attack (PAA)}

Typically, neural solvers decompose an MOCOP into a series of subproblems with different preferences, which are solved independently. According to \cite{lin2022pareto}, if a neural solver can solve subproblems well with any preference $\lambda$, it can generate a good approximation to the whole Pareto front for MOCOP. 
In this paper, we hypothesize that if a neural model does not effectively approximate the solution of the subproblem under certain values of $\lambda$, the resulting approximation of the Pareto front will be inadequate. Following this inspiration, we propose the PAA method to attack neural solvers for MOCOPs.
In particular, perturbations are applied to clean instances under different preferences, resulting in hard instances tailored to each preference. 

For each preference-conditioned subproblem, we construct hard instances by maximizing a variant of the reinforcement loss, defined as:
\begin{equation}
\ell(x; \theta) = \frac{L(\pi \mid x)}{b(x)} \log p_\theta(\pi \mid x),
\label{eq:loss}
\end{equation}
where \(L(\pi \mid x)\) represents the loss of the subproblem with a given preference \(\lambda\).
\(b(x)\) is the baseline of \(L(\pi \mid x)\), which is calculated by \(b(x) = \frac{1}{M} \sum_{j=1}^M L(\pi_j \mid x)\), where \(M\) is the number of sampled tours (i.e., feasible solutions) for the given instance.
$x$ denotes the problem-specific input, i.e., node coordinates in TSP.
$p_\theta(\pi \mid x)$ denotes the probability distribution of the solution \(\pi\), which is derived from a neural model parameterized by \(\theta\). 

The input \(x\) corresponding to each preference undergoes the following iterative update:

\begin{equation}
x^{(t+1)} = \Pi_{\mathcal{N}} \left[ x^{(t)} + \alpha \cdot \nabla_{x^{(t)}} \ell(x^{(t)}; \theta^{(t)}) \right],
\label{eq:gradient}
\end{equation}
where \(\theta^{(t)}\) denotes the best-performing model in iteration \(t\), \(\mathcal{N}\) represents the feasible solution space, \(\alpha\) is the step size and \(\ell(x^{(t)}; \theta^{(t)})\) is the reinforcement loss defined in Eq.(\ref{eq:loss}). At each iteration \(t\), the variable \(x^{(t)}\) is updated by making a gradient ascent on the loss function \(\ell(x; \theta)\), with the calculated gradients \(\nabla_{x^{(t)}} \ell(x^{(t)}; \theta^{(t)})\) guiding the update step. 
The projection operator $\Pi_{\mathcal{N}}(\cdot)$ is a min-max normalization, ensuring that the updated variables $x^{(t+1)}$ remain within a feasible solution space $\mathcal{N}$.
The iterative process continues until the variable \(x^{(t)}\) converges towards hard instances for the current given preference.
These instances are then aggregated into an adversarial instance pool for robustness evaluation and the subsequent defense stage.

\subsection{Dynamic Preference-augmented Defense (DPD)}
To improve robustness, DPD performs adversarial fine-tuning on a mixture of clean instances and a pre-generated adversarial instance pool. 
Specifically, we first construct an adversarial pool $\mathcal{D}_{\mathrm{hard}}$ by applying PAA to clean instances under a predefined preference set $\Lambda$ using the pretrained solver $\theta_0$. 
This pool is kept fixed during DPD training. 
At each training epoch, we newly sample clean instances $\mathcal{D}_{\mathrm{clean}}^{t}$ and mix them with $\mathcal{D}_{\mathrm{hard}}$ using balanced sampling to form the training data:
\begin{equation}
\mathcal{D}_{\mathrm{train}}^{t}
=
\mathcal{D}_{\mathrm{clean}}^{t}
\cup
\mathcal{D}_{\mathrm{hard}} .
\end{equation}

For each mini-batch, we sample instance--preference pairs $(x,\lambda)$ from $\mathcal{D}_{\mathrm{train}}^{t}$, where $\lambda = (\lambda_1,\lambda_2,\dots,\lambda_m)$ is a preference vector. 
For each sampled preference $\lambda$, we generate a set of locally augmented preferences $\{\lambda_1', \lambda_2', \dots, \lambda_N'\}$ to explore its neighborhood in the preference space. 
The purpose of this augmentation is to identify nearby preference regions where the current model performs poorly. 
For each preference vector $\lambda$, its augmented preferences are computed as:

\begin{equation}
\lambda_i' = \text{Perturb}(\lambda, \delta_i),
\label{eq:7}
\end{equation}
where \(\delta_i \sim \text{Uniform}(-\epsilon, \epsilon)\) is a small random perturbation. \(i\) reflects the index of the perturbed preference vector.
Since the perturbation may result in a preference vector that does not satisfy the constraint \(\sum_{k=1}^m \lambda_{i,k}' = 1\), a normalization step is applied to ensure validity:
\begin{equation}
\lambda_{i,k}' = \frac{\lambda_{i,k}'^{\text{raw}}}{\sum_{j=1}^m \lambda_{i,j}'^{\text{raw}}}, \quad \forall k \in \{1, \dots, m\},
\label{eq:8}
\end{equation}
where \(\lambda_{i,k}'^{\text{raw}}\) represents the raw preference value after perturbation. By incorporating the normalization step, the generated preferences remain within the valid preference space, ensuring \(\sum_{k=1}^m \lambda_{i,k}' = 1\) for all augmented preferences.
\begin{algorithm}[t]
\caption{\textbf{DPD with a Pre-generated Adversarial Pool}}\small
\label{alg:paa_dpd}
\footnotesize
\raggedright
\textbf{Input:} 
pre-trained model $\theta_0$, preference set $\Lambda = \{\lambda_k\}_{k=1}^P$,
epochs $E$, batch size $B$, 
number of perturbed preferences $N$, optimizer $\text{ADAM}$, number of mini-batches per epoch $M$

\textbf{Output:} Updated model parameters $\theta$

\begin{algorithmic}[1]
\State Initialize $\theta \leftarrow \theta_0$.
\State Generate seed clean instances $\mathcal{D}_{\mathrm{seed}}$ with the uniform distribution.
\State $\mathcal{D}_{\mathrm{hard}} \leftarrow \emptyset$.

\For{$k = 1$ to $P$}
    \State Select preference $\lambda_k \in \Lambda$.
    \State Generate $d^{\mathrm{adv},k}$ from $\mathcal{D}_{\mathrm{seed}}$ using preference $\lambda_k$ via PAA with model $\theta_0$.
    \State $\mathcal{D}_{\mathrm{hard}} \leftarrow \mathcal{D}_{\mathrm{hard}} \cup d^{\mathrm{adv},k}$.
\EndFor

\For{$t = 1$ to $E$}
    \State Generate clean instances $\mathcal{D}_{\mathrm{clean}}^{t}$ with the uniform distribution.
    \State $\mathcal{D}_{\mathrm{train}}^{t} \leftarrow \mathcal{D}_{\mathrm{clean}}^{t} \cup \mathcal{D}_{\mathrm{hard}}$.

    \For{$j = 1$ to $M$}
        \State Sample a balanced mini-batch $\mathcal{B}$ of size $B$ from $\mathcal{D}_{\mathrm{clean}}^{t}$ and $\mathcal{D}_{\mathrm{hard}}$.
        \State Sample a preference $\lambda_j \in \Lambda$ for $\mathcal{B}$.

        \For{$i = 1$ to $N$}
            \State Generate $\lambda_i'$ using Eq.~(\ref{eq:7}) and Eq.~(\ref{eq:8}).
            \State Estimate $Tch(\lambda_i')$ on $\mathcal{B}$ with model parameters fixed.
        \EndFor

        \State Select $\lambda_{\mathrm{adv}}' = \arg\max_i Tch(\lambda_i')$ according to Eq.~(\ref{eq:10}).
        \State Compute gradient $\nabla \mathcal{J}(\theta)$ using $\lambda_{\mathrm{adv}}'$ and $\mathcal{B}$.
        \State Update parameters: $\theta \gets \text{ADAM}(\theta, \nabla \mathcal{J}(\theta))$.
    \EndFor
\EndFor
\end{algorithmic}
\end{algorithm}

For each augmented preference $\lambda_i'$, the neural solver performs a forward inference on the current mini-batch to generate solutions and compute the corresponding Tchebycheff values, while keeping the model parameters fixed.

The value $Tch(\lambda_i')$ quantifies the quality of the solutions generated by the model under preference $\lambda_i'$, where a higher value $Tch(\lambda_i')$ indicates a inferior quality of the solution for this preference. 
Since the number of augmented preferences $N$  is small, the additional cost of preference search remains limited in practice. These augmented preferences can be evaluated in parallel within a mini-batch, resulting in only a modest increase in training overhead.
The preference that yields the largest Tchebycheff value is selected for further optimization, i.e.,
\begin{equation}
\lambda_{\text{adv}}' = \arg\max_{i} Tch(\lambda_i').
\label{eq:10}
\end{equation}

For the selected adversarial preference $\lambda_{\mathrm{adv}}'$, 
we update the solver on the mixed mini-batch $\mathcal{B}$ sampled from 
$\mathcal{D}_{\mathrm{train}}^{t}$. 
For each instance $x \in \mathcal{B}$, the scalarized loss is computed as:
\begin{equation}
L(x \mid \lambda_{\text{adv}}') = \max_{1 \leq k \leq m} \lambda_{\text{adv},k}' \cdot |f_k(x) - z_k^*|.
\label{eq:11}
\end{equation}

The optimization process uses the REINFORCE algorithm \cite{williams1992simple} to minimize loss.The proposed framework is implemented through the training procedure summarized in Algorithm~\ref{alg:paa_dpd}.
\section{Experiments}

In this section, we conduct a comprehensive set of experiments on four MOCOPs: bi-objective TSP (Bi-TSP), tri-objective TSP (Tri-TSP), bi-objective CVRP (Bi-CVRP), and bi-objective KP (Bi-KP)  to thoroughly analyze and evaluate the effectiveness of the proposed attack and defense methods. All experiments are executed on a server equipped with an Intel(R) Xeon(R) Silver 4214R CPU @ 2.40GHz and an RTX 3090 GPU.

\subsection{Baselines and Settings}

\noindent \textbf{Instance Distributions for Evaluation.}
To evaluate the efficacy of the proposed attack approach, we benchmark against four typical instance distributions (clean uniform, log-normal (0,1) with moderate skewness, beta (2,5) with bounded asymmetry, and gamma (2,0.5) with high skewness) as well as ROCO \cite{lu2023roco}, a learning-based attack method that perturbs graph edges under a no-worse-optimum guarantee and trains an agent with PPO \cite{schulman2017ppo} to maximize solver degradation.

\noindent \textbf{Evaluation Setup for Targeted Solvers.}
We target representative neural MOCOP solvers, namely Conditional Neural Heuristic \textbf{CNH}, Meta Neural Heuristic \textbf{EMNH} ~\cite{chen2024efficient}, and Preference-Based Neural Heuristic \textbf{PMOCO}. We selected these solvers as they all adopt \textbf{POMO} ~\cite{kwon2020pomo} as the base model for solving single-objective subproblems. For fair comparisons, we adopt WS (weighted sum) scalarization across all methods.
To establish the baseline for the relative optimality gap, we approximate the Pareto front using two non-learnable solvers: \textbf{WS-LKH} for MOTSP and MOCVRP, and weighted-sum dynamic programming \textbf{(WS-DP)} for MOKP.

\noindent \textbf{Metrics.}
To evaluate the proposed attack and defense methods, the average \textbf{HV} \cite{audet2021performance} and the average optimality \textbf{gap} are employed. HV provides a comprehensive measure of both the diversity and convergence of solutions, while the gap quantifies the relative difference in HV compared to the first baseline solver. 


\noindent \textbf{Implementations.}
We evaluate PMOCO, CNH, and EMNH using their pre-trained models. Hard instances are generated with 101 and 105 uniformly sampled preferences for the bi- and tri-objective settings, respectively, with 100 clean samples per preference, yielding 10,100 and 10,500 instances. Training uses 10,000 clean samples plus hard instances per epoch, with 3 gradient steps (step size 0.01) over 200 epochs. For testing, 200 Gaussian instances are constructed with $c_{\text{DIST}} \in [10,20,30,40,50]$. Other settings (e.g., learning rate, batch size) follow their original papers.

\begin{table*}[!t]
\centering
\caption{Attack performance on 100-node instances. Bold values indicate the strongest attack (lowest HV / largest Gap) for each solver.}
\vspace{-2mm}
\setlength{\tabcolsep}{4pt}
\renewcommand{\arraystretch}{0.9}
\label{tab:attack_main}
\begin{threeparttable}
\resizebox{\textwidth}{!}{
\begin{tabular}{ll|cc|cc|cc|cc|cc|cc}
\toprule
\multirow{2}{*}{\textbf{Problem}} 
& \multirow{2}{*}{\textbf{Method}} 
& \multicolumn{2}{c|}{\textbf{Clean}} 
& \multicolumn{2}{c|}{\textbf{LogNormal(0,1)}} 
& \multicolumn{2}{c|}{\textbf{Beta(2,5)}} 
& \multicolumn{2}{c|}{\textbf{Gamma(2,0.5)}} 
& \multicolumn{2}{c|}{\textbf{ROCO-RL}} 
& \multicolumn{2}{c}{\textbf{PAA}} \\
\cmidrule(lr){3-4} \cmidrule(lr){5-6} \cmidrule(lr){7-8} \cmidrule(lr){9-10} \cmidrule(lr){11-12} \cmidrule(lr){13-14}
& 
& \textbf{HV} $\downarrow$ & \textbf{Gap} $\uparrow$
& \textbf{HV} $\downarrow$ & \textbf{Gap} $\uparrow$
& \textbf{HV} $\downarrow$ & \textbf{Gap} $\uparrow$
& \textbf{HV} $\downarrow$ & \textbf{Gap} $\uparrow$
& \textbf{HV} $\downarrow$ & \textbf{Gap} $\uparrow$
& \textbf{HV} $\downarrow$ & \textbf{Gap} $\uparrow$ \\
\midrule

\multirow{4}{*}{\textbf{Bi-TSP100}}
& \textbf{WS-LKH} & 0.6799 & -- & 0.9123 & -- & 0.7201 & -- & 0.7771 & -- & 0.7094 & -- & 0.6824 & -- \\
& \textbf{EMNH}   & 0.6653 & 2.15\% & 0.9034 & 0.97\% & 0.7119 & 1.14\% & 0.7726 & 0.58\% & 0.6914 & \textbf{2.53\%} & 0.6665 & 2.33\% \\
& \textbf{PMOCO}  & 0.6571 & \textbf{3.34\%} & 0.8948 & 1.92\% & 0.7035 & 2.33\% & 0.7705 & 0.84\% & 0.6946 & 2.09\% & 0.6603 & 3.23\% \\
& \textbf{CNH}    & 0.6682 & 1.64\% & 0.9049 & 0.81\% & 0.7130 & 0.98\% & 0.7741 & 0.38\% & 0.6957 & 1.93\% & 0.6621 & \textbf{2.12\%} \\
\midrule

\multirow{4}{*}{\textbf{Bi-CVRP100}}
& \textbf{WS-LKH} & 0.2408 & -- & 0.7265 & -- & 0.7435 & -- & 0.7252 & -- & 0.3782 & -- & 0.2705 & -- \\
& \textbf{EMNH}   & 0.2309 & 4.11\% & 0.7240 & 0.34\% & 0.7363 & 0.97\% & 0.7149 & 0.04\% & 0.3709 & 1.93\% & 0.2588 & \textbf{4.32\%} \\
& \textbf{PMOCO}  & 0.2307 & 4.19\% & 0.7248 & 0.23\% & 0.7375 & 0.81\% & 0.7050 & 2.77\% & 0.3703 & 2.08\% & 0.2591 & \textbf{4.21\%} \\
& \textbf{CNH}    & 0.2393 & 0.62\% & 0.7235 & 0.41\% & 0.7389 & 0.62\% & 0.7179 & 1.00\% & 0.3718 & 1.69\% & 0.2597 & \textbf{3.99\%} \\
\midrule

\multirow{4}{*}{\textbf{Bi-KP100}}
& \textbf{WS-DP}  & 0.8283 & -- & 0.6628 & -- & 0.6046 & -- & 0.8012 & -- & -- & -- & 0.6483 & -- \\
& \textbf{EMNH}   & 0.8571 & -3.47\% & 0.6827 & -3.00\% & 0.6189 & -2.36\% & 0.7984 & 0.35\% & -- & -- & 0.6056 & \textbf{6.59\%} \\
& \textbf{PMOCO}  & 0.8549 & -3.22\% & 0.7058 & -6.49\% & 0.6241 & -3.21\% & 0.8181 & -2.11\% & -- & -- & 0.6023 & \textbf{7.09\%} \\
& \textbf{CNH}    & 0.8608 & -3.93\% & 0.7028 & -6.04\% & 0.6239 & -3.19\% & 0.8194 & -2.27\% & -- & -- & 0.6094 & \textbf{6.00\%} \\
\midrule

\multirow{4}{*}{\textbf{Tri-TSP100}}
& \textbf{WS-LKH} & 0.4599 & -- & 0.8663 & -- & 0.6392 & -- & 0.7392 & -- & 0.4874 & -- & 0.4490 & -- \\
& \textbf{EMNH}   & 0.4412 & 4.07\% & 0.8309 & 4.08\% & 0.6114 & 4.35\% & 0.7118 & 3.71\% & 0.4659 & \textbf{4.41\%} & 0.4296 & 4.32\% \\
& \textbf{PMOCO}  & 0.4349 & 5.42\% & 0.8324 & 3.92\% & 0.6141 & 3.93\% & 0.7112 & 3.78\% & 0.4620 & 5.21\% & 0.4237 & \textbf{5.63\%} \\
& \textbf{CNH}    & 0.4401 & 4.31\% & 0.8417 & 2.84\% & 0.6180 & 3.31\% & 0.7164 & 3.08\% & 0.4641 & 4.77\% & 0.4253 & \textbf{5.27\%} \\
\bottomrule
\end{tabular}
}
\end{threeparttable}
\end{table*}
\subsection{Attack Performance}
From Table ~\ref{tab:attack_main}, it can be observed that perturbations based on log-normal, beta, and gamma distributions generally have little  effect on reducing the HV value of the solution set. 
In particular, these perturbations produce higher HV values across various solvers compared to clean instances. This indicates that conventional disturbances struggle to substantially impair the performance of solvers such as WS-LKH and WS-DP. Furthermore, the discrepancies between the solutions generated by these neural solvers and the conventional solver under these distributions are consistently smaller than those observed for clean instances. Hence, despite the heterogeneous nature of these distributions, neural solvers demonstrate robust capabilities to maintain high-quality solutions.
In contrast, PAA generates problem distributions that significantly reduce HV values in both classical and neural MOCOP solvers, demonstrating strong and consistent attack effect across all problems and sizes. Notably, it achieves the best attack effect over all cases in Bi-KP.

Furthermore, the HV gaps of different solvers on hard instances generated by PAA and on clean instances are considerably larger. For example, on Bi-CVRP100, the attack against EMNH yields a gap of 4.32\%, while on Bi-KP100, the attack against PMOCO reaches 7.09\%, significantly exceeding the attack effects by instances generated by the other methods.
ROCO-RL shows non-trivial attack capability on a few instances (e.g., a notable 4.41\% gap against EMNH on Tri-TSP100), yet PAA consistently surpasses it in most MOCO problems, achieving superior attack performance.
This indicates that PAA explicitly exposes the vulnerabilities of diverse neural MOCOP solvers, underscoring its effectiveness.


\subsection{Defense Performance}

To evaluate our defense method, we conducted comparative experiments on EMNH, PMOCO, and CNH trained on uniformly distributed clean instances, alongside their DPD variants trained under the proposed framework. We also include WE-CA~\cite{chen2025rethinking}, a recent neural MOCOP solver that adopts feature-wise affine transformations in the encoder to enhance preference-conditioned representations. All models (with or without DPD) were evaluated in Gaussian instances. The results are reported in Table~\ref{tab:defense}.
As shown, DPD-defended solvers (PMOCO-DPD, CNH-DPD, EMNH-DPD, WE-CA-DPD) consistently enhance the performance of neural solvers, achieving overall improvements on all problems. Remarkably, on Bi-TSP20 and Bi-CVRP100, WE-CA-DPD and CNH-DPD achieve the first and second best results, respectively. The improvement is particularly evident on Bi-CVRP100, where WE-CA-DPD improves HV by 2.23\% over WS-LKH, the largest gain among all solvers.

\begin{table}[!t]
\centering
\caption{Optimality Gap Analysis for Defense Performance. \textbf{Bold} values indicate the best performance in the respective metric. \underline{Underlined} values indicate the second-best performance in the respective metric.}
\vspace{-3mm}
\setlength{\tabcolsep}{4pt}
\renewcommand{\arraystretch}{0.85} 
\label{tab:defense}
\begin{threeparttable}
\resizebox{\linewidth}{!}{
\begin{tabular}{l|c|ccc|ccc|ccc}
\toprule
\multirow{2}{*}{Method} & \multirow{2}{*}{Instance} & \multicolumn{3}{c|}{20 Nodes} & \multicolumn{3}{c|}{50 Nodes} & \multicolumn{3}{c}{100 Nodes} \\
                        &                          & HV $(\uparrow)$     & Gap $(\downarrow)$  & Time $(\downarrow)$ & HV $(\uparrow)$    & Gap $(\downarrow)$ & Time$(\downarrow)$  & HV $(\uparrow)$    & Gap $(\downarrow)$ & Time $(\downarrow)$ \\ \midrule
WS-LKH & \multirow{9}{*}{Bi-TSP}  & 0.8873  & - & 4.30m  & 0.8660   & -  & 38.47m  & 0.8365   & - & 3.19h \\
EMNH           &                          & 0.8742  & 1.48\% & 6.12s & 0.8649   & 0.13\%  & 9.42s & 0.8265   & 1.19\%  & 30.09s \\
\textbf{EMNH-DPD}       &                          & 0.8894 & -0.23\% & 6.38s & \textbf{0.8697} & \textbf{-0.43\%}  & 9.15s & 0.8317   & 0.57\% & 30.25s \\
PMOCO          &                          & 0.8779  & 1.06\% & 7.11s & 0.8566   & 1.08\%  & 11.33s & 0.8248   & 1.40\%  & 30.87s \\
\textbf{PMOCO-DPD}      &                          & 0.8867  & 0.07\% & 8.44s & 0.8654   & 0.06\%  & 13.29s & 0.8360   & 0.06\% & 32.08s \\
CNH            &                          & 0.8794  & 0.89\% & 7.32s & 0.8587   & 0.84\%  & 12.03s & 0.8294   & 0.85\%  & 34.76s \\
\textbf{CNH-DPD}        &                          & \underline{0.8871}  & \underline{0.02\%} & 8.67s & 0.8660   & 0.00\%  & 15.22s & \underline{0.8373} & \underline{-0.09\%} & 33.57s \\ 
WE-CA        &                          & 0.8803  & 0.78\% & 7.33s &  0.8591  & 0.79\%  & 10.52s &  0.8304  & 0.72\% & 31.34s \\ 
\textbf{WE-CA-DPD}        &                          & \textbf{0.8886}  & \textbf{-0.14\%} &  7.29s &   \underline{0.8674} & \underline{-0.16\%}  & 10.17s & \textbf{0.8377}  & \textbf{-0.14\%} & 31.41s  \\ 
\midrule
WS-LKH        & \multirow{9}{*}{Bi-CVRP} & 0.5743  & - & 6.44m & 0.5314   & - & 44.82m & 0.5157   & - & 4.03h \\
EMNH           &                          & 0.5558  & 3.22\% & 6.03s & 0.5278   & 0.67\%  & 16.11s & 0.5048   & 2.11\%  & 40.29s \\
\textbf{EMNH-DPD}       &                          & 0.5772  & -0.50\% & 6.72s & 0.5319   & -0.09\%  & 16.88s & 0.5258   & -1.96\% & 40.75s \\
PMOCO          &                          & 0.5526  & 3.78\% & 6.39s & 0.5219   & 1.79\%  & 18.02s & 0.5015   & 2.75\%  & 47.21s \\
\textbf{PMOCO-DPD}      &                          & 0.5763  & -0.35\% & 6.51s & 0.5308   & 0.11\%  & 17.44s & 0.5173   & -0.31\% & 47.93s \\
CNH            &                          & 0.5564  & 3.11\% & 7.23s & 0.5289   & 0.47\%  & 19.55s & 0.5071   & 1.67\%  & 52.16s \\
\textbf{CNH-DPD}        &                          & \underline{0.5794} & \underline{-0.88\%} & 7.65s & \textbf{0.5386} & \textbf{-1.35\%}  & 19.98s & \underline{0.5261} & \underline{-2.01\%} & 51.49s \\ 
WE-CA        &                          & 0.5572  & 2.97\% & 6.41s &  0.5292  & 0.41\%  & 16.43s & 0.5109   & 0.93\% & 44.24s \\ 
\textbf{WE-CA-DPD}        &                         &  \textbf{0.5803}  & \textbf{-1.04\%} & 6.23s & \underline{0.5382}   & \underline{-1.27\%}  & 16.37s &  \textbf{0.5272}  & \textbf{-2.23\%} & 43.72s \\ 

\midrule
WS-DP  & \multirow{9}{*}{Bi-KP}   & \textbf{0.5832} & - & 17.45m & \textbf{0.4948} & - & 1.42h & \textbf{0.6783} & - & 4.23h \\ 
EMNH           &                          & 0.5817  & 0.26\% & 5.32s & 0.4858   & 1.81\%  & 17.46s & 0.6682   & 1.49\%  & 40.23s \\
\textbf{EMNH-DPD}       &                          & 0.5828  & 0.06\% & 5.38s & 0.4903   & 0.91\%  & 18.45s & 0.6718   & 0.96\% & 40.74s \\
PMOCO         &                          & 0.5809  & 0.39\% & 7.22s & 0.4803   & 2.93\%  & 16.87s & 0.6653   & 1.91\%  & 48.31s \\
\textbf{PMOCO-DPD}      &                          & \underline{0.5829}  & \underline{0.05\%} & 7.88s & 0.4897   & 1.03\%  & 17.94s & 0.6712   & 1.04\% & 48.92s \\
CNH            &                          & 0.5820  & 0.21\% & 8.17s & 0.4845   & 2.08\%  & 19.12s & 0.6683   & 1.47\%  & 54.11s \\
\textbf{CNH-DPD}        &                          & \textbf{0.5832} & \textbf{0.00\%} & 8.49s & 0.4901   & 0.95\%  & 19.46s & 0.6742   & 0.60\% & 53.74s \\ 
WE-CA        &                          & 0.5823 & 0.15\% & 7.24s & 0.4852   & 1.94\%  & 16.18s &  0.6691  & 1.35\% & 45.72s \\ 
\textbf{WE-CA-DPD}        &                          & \underline{ 0.5829 }  & \underline{ 0.05\%} &  7.53s &  \underline{0.4913}  & \underline{0.70\%}  & 17.44s &  \underline{0.6752}  & \underline{0.46\%} & 44.43s \\ 

\midrule

WS-LKH & \multirow{9}{*}{Tri-TSP} & 0.6864  & - & 6.03m & \textbf{0.6151} & - & 55.14m & \textbf{0.4978} & - & 3.71h \\
EMNH           &                          & 0.6719  & 2.11\% & 5.12s & 0.5831   & 5.20\%  & 9.43s & 0.4829   & 2.99\%  & 30.31s \\ 
\textbf{EMNH-DPD}       &                          & \textbf{0.6885} & \textbf{-0.31\%} & 5.29s & \underline{0.6144}   & \underline{0.11\%}  & 9.67s & \underline{0.4960}   & \underline{0.36\%} & 30.52s \\ 
PMOCO          &                          & 0.6708  & 2.27\% & 6.18s & 0.5954   & 3.20\%  & 11.22s & 0.4825   & 3.07\%  & 30.32s \\
\textbf{PMOCO-DPD}      &                          & 0.6817  & 0.68\% & 7.08s & 0.6079   & 1.17\%  & 12.34s & 0.4930   & 0.96\% & 32.21s \\
CNH            &                          & 0.6791  & 1.06\% & 7.32s & 0.6049   & 1.65\%  & 12.67s & 0.4872   & 2.13\%  & 33.86s \\
\textbf{CNH-DPD}        &                          & \underline{0.6877}  & \underline{-0.18\%} & 8.11s & 0.6127   & 0.39\%  & 15.42s & 0.4938   & 0.80\% & 33.14s \\ 
WE-CA        &                          & 0.6793  & 1.03\% &  6.33s &  0.6051  & 1.63\%  &  12.45s & 0.4876  & 2.04\% & 31.41s \\ 
\textbf{WE-CA-DPD}        &                          & 0.6862  &  0.03\% & 7.43s & 0.6133   & 0.29\%  & 13.09s &  0.4952  & 0.52\% & 31.49s \\ 

\bottomrule
\end{tabular}
}
\end{threeparttable}
\end{table}
In addition, CNH-DPD achieves the best result on Bi-CVRP50. The strong and consistent Bi-CVRP results indicate that models with encoder-level preference-instance interaction mechanism (e.g., CNH, WE-CA) exhibit the most pronounced improvements under DPD. In particular, CNH-DPD and WE-CA-DPD deliver leading performance on Bi-CVRP20/50/100.

Regarding the meta-learning–based solver, EMNH-DPD improves EMNH performance and produces the best result (HV 0.6885, with a runtime of 5.29s) on Tri-TSP20, as well as second-best results on Tri-TSP50 and Tri-TSP100. This demonstrates the versatility of DPD in enhancing solvers across distinct learning paradigms.

In terms of computational efficiency, DPD-defended solvers require considerably less runtime compared to non-learnable solvers. For example, WS-DP requires 17.45 minutes to reach the best HV value on Bi-KP, while CNH-DPD in only 8.49 seconds achieves the same. Overall, these results demonstrate that DPD substantially enhances the robustness of neural solvers, yielding strong generalization to larger problem sizes and distribution shifts.

\subsection{Ablation Study}
Ablation studies were conducted on critical hyperparameters of the proposed attack method, with experiments performed on three-objective 50-node TSP instances.
\paragraph{Impact of Gradient Iteration Counts}
\begin{figure}[!t]
    \centering
    \includegraphics[width=0.8\linewidth]{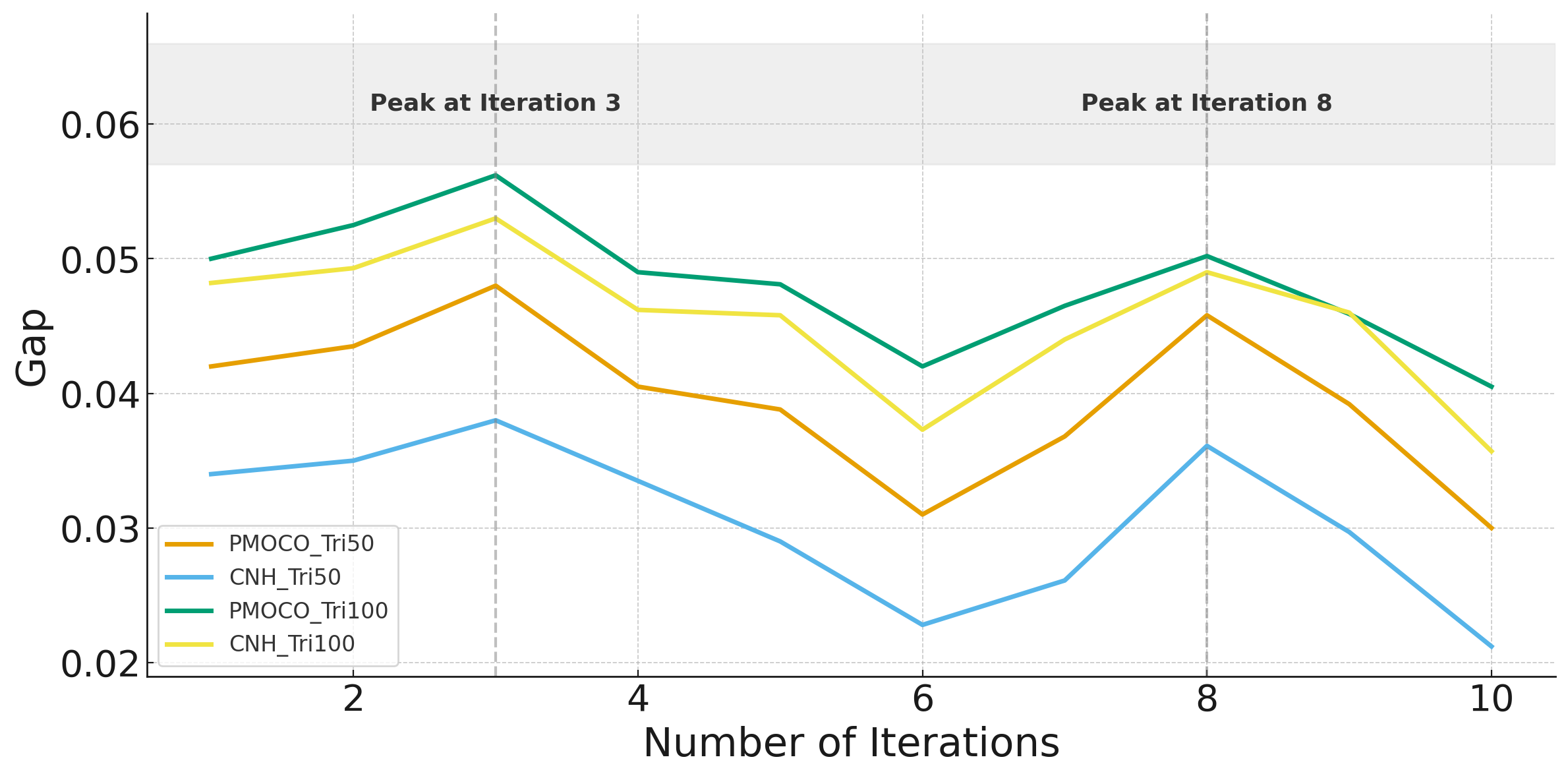}
    \caption{Impact of Iteration Counts on HV and Gap.}
    \label{fig:iteration}
\end{figure}
The iteration count \(t\) in Eq.~(\ref{eq:gradient}) of the main paper is varied from 1 to 10 to evaluate its impact on the HV values and the gap relative to the LKH. As illustrated in Figure ~\ref{fig:iteration}, the gap peaks at \(t = 3\) and \(t = 8\), with the maximum observed at \(t = 3\). Consequently, \(t = 3\) is selected in our experiment to balance computational efficiency and performance analysis.
\begin{figure}[ht]
    \centering
    \includegraphics[width=0.65\textwidth]{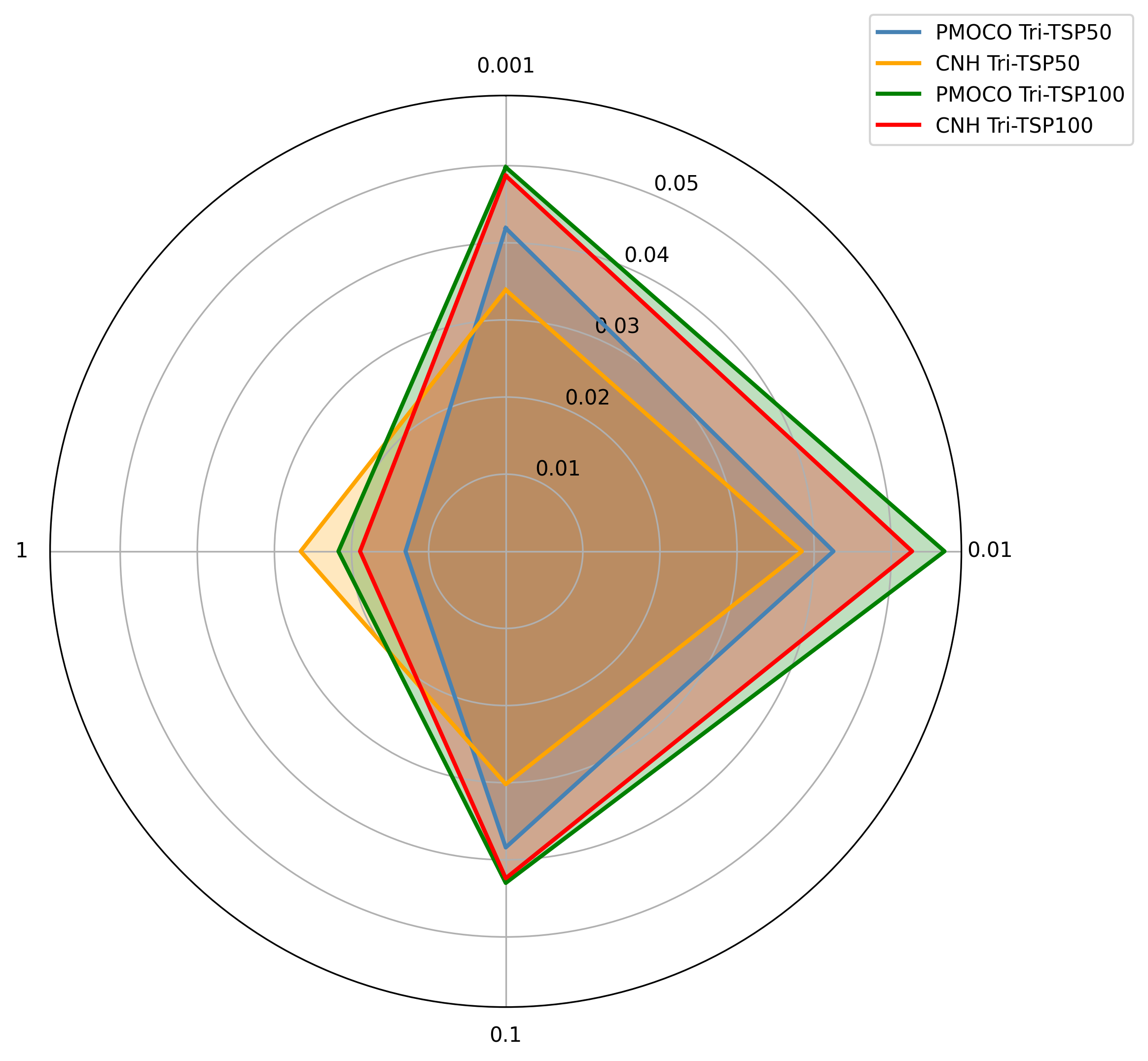}
    \caption{HV Gaps for Different $\alpha$.}
    \label{fig:gradient updata}
\end{figure}
\paragraph{Impact of Gradient Update Parameters}
Figure~\ref{fig:gradient updata} (radar chart) illustrates the relationship between the step size $\alpha$ in Eq.~(\ref{eq:gradient}) of the main paper and their HV gaps, showing that the gap reaches its maximum values in $\alpha = 0.01$. Therefore, to maximize the effectiveness of the attack, $\alpha = 0.01$ is adopted in our experiments.

\paragraph{Benchmark Evaluations}
Similarly to previous studies~\cite{fan2024conditional}~\cite{9040280}, we evaluated the performance of our DPD framework on six Bi-TSP100 benchmark instances\footnote{\url{https://sites.google.com/site/kflorios/motsp?pli=1}}: kroAB100, kroAC100, kroAD100, kroBC100, kroBD100 and kroCD100, which were constructed by combining instances from the kroA100, kroB100, kroC100, and kroD100 instances.
\begin{figure}[!t]
    \centering
    \includegraphics[width=0.6\linewidth]{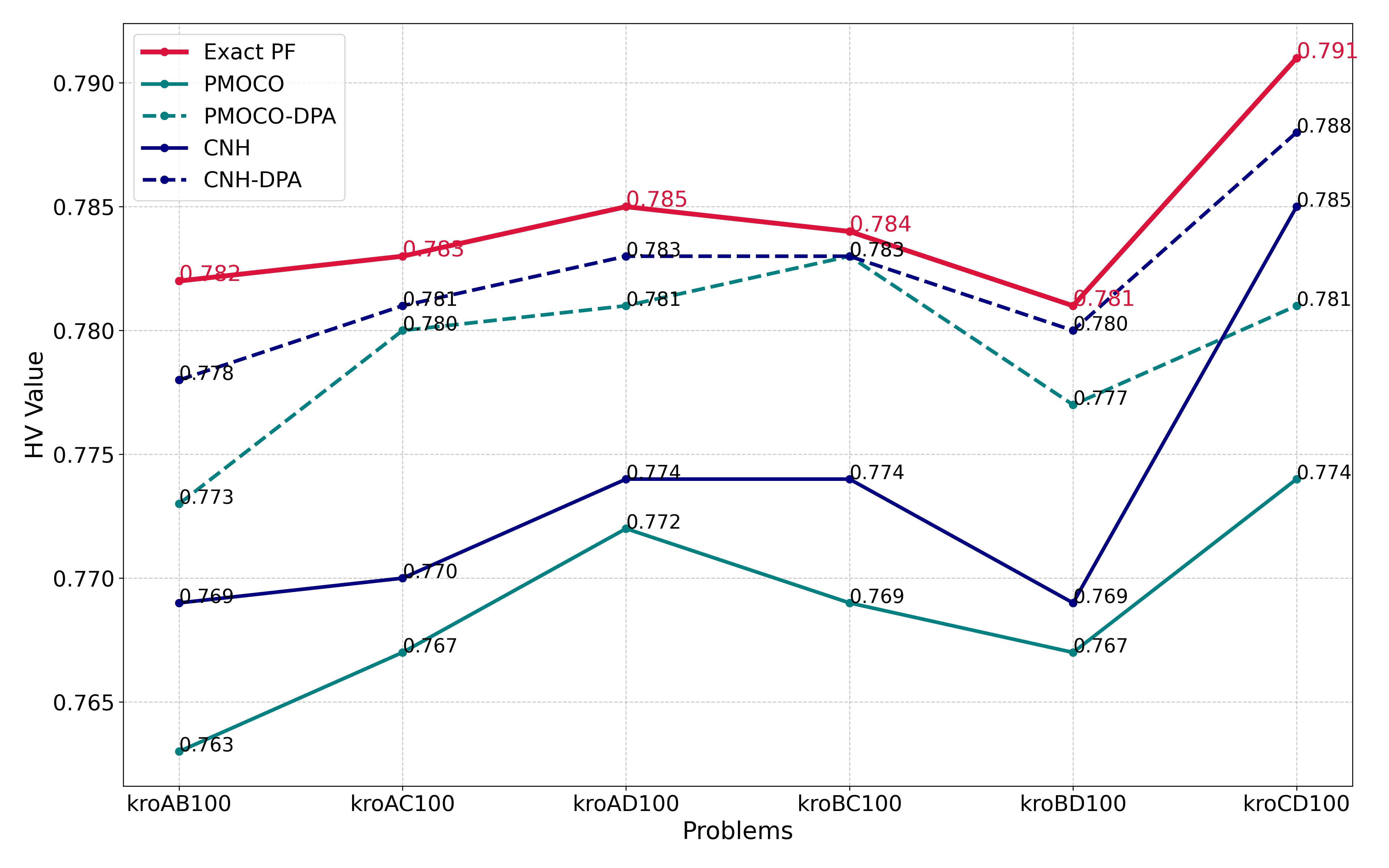}
    \caption{Benchmark Performance Comparison on HV Metric.}
    \label{fig:Benchmark}
\end{figure}
As illustrated in Figure ~\ref{fig:Benchmark}, models trained on the hard instances consistently outperform those trained on the clean instances in all the problem instances. The CNH-DPD model achieves the HV values among the learned models, closely approaching the exact PF. In particular, in kroAC100, PMOCO-DPD and CNH-DPD achieve HV values that are 1.4\% and 1.7\% higher than those of PMOCO and CNH, respectively. In kroBC100 and kroBD100, the HV values for DPD-enhanced models are within 0.1\% of the exact PF, demonstrating their competitive performance and robustness. These results underscore the effectiveness of the proposed approach in handling diverse instance distributions and enhancing solver adaptability under adversarial conditions.

\paragraph{Generalization Study}
We evaluate the generalization capability of DPD on two types of larger scale test instances (\(n = 150/200\)) including clean instances and mixed Gaussian instances. As illustrated in Table ~\ref{table:large}, our model demonstrates remarkable robustness across both test scenarios while maintaining strong performance under varying instance distributions.
\begin{table}[!t]
\centering
\caption{Comparison of Bi-TSP performance with $n=150$ and $n=200$ on 200 clean and Mix Gaussian test instances.}
\vspace{-3mm}
\setlength{\tabcolsep}{4pt}
\renewcommand{\arraystretch}{0.85} 
\resizebox{\textwidth}{!}{ 
\begin{tabular}{l|ccc|ccc|ccc|ccc}
\hline
\label{table:large}
\multirow{2}{*}{Method} & \multicolumn{6}{c|}{Clean Instances} & \multicolumn{6}{c}{Gaussian Instances} \\
                        & \multicolumn{3}{c|}{Bi-TSP ($n=150$)} & \multicolumn{3}{c|}{Bi-TSP ($n=200$)} & \multicolumn{3}{c|}{Bi-TSP ($n=150$)} & \multicolumn{3}{c}{Bi-TSP ($n=200$)} \\
                        & HV  & Gap & Time   & HV  & Gap & Time    & HV  & Gap & Time   & HV  & Gap & Time    \\ \hline
WS-LKH & \textbf{0.7149} & - & 13h    & \textbf{0.7490} & - & 22h     & \textbf{0.8506} & - & 13h    & \textbf{0.8790} & - & 22h     \\
PMOCO          & 0.7028         & 1.69\%          & 55.38s    & 0.7318         & 2.29\%          & 1.52m      & 0.8367         & 1.63\%          & 55.87s    & 0.8608         & 2.07\%          & 1.52m     \\
\textbf{PMOCO-DPD}      & 0.7091         & 0.81\%          & 57.22s    & 0.7327         & 2.17\%          & 1.59m     & \underline{0.8430}         & \underline{0.89\%}          & 57.22s    & \underline{0.8660}         & \underline{1.47\%}          & 1.59m     \\
CNH           & 0.7043         & 1.48\%          & 57.45s    & 0.7324         & 2.21\%          & 1.53m    & 0.8379         & 1.49\%          & 57.49s    & 0.8598         & 2.18\%          & 1.53m    \\
\textbf{CNH-DPD}        & \underline{0.7104}         & \underline{0.63\%}          & 58.33s    & \underline{0.7374}         & \underline{1.54\%}          & 2.02m    & 0.8427         & 0.92\%          & 58.36s    & 0.8649         & 1.60\%          & 2.05m    \\ \hline
\end{tabular}
}
\end{table}

\section{Conclusions}
In this paper, we investigate the robustness and performance of state-of-the-art neural MOCOP solvers under diverse hard and clean instances distributions. We proposed an innovative attack method that effectively generates hard (challenging) problem instances, measuring the vulnerability in solver's performance by reducing HV values and increasing optimality gaps compared to baseline methods. Furthermore, we also proposed a defense method that leverages adversarial training with hardness-aware preference selection, showing improved robustness across various solvers and tasks. These two methods contribute to solving multi-objective optimization challenges by enhancing the robustness and generalizability of neural solvers, leading to more robust solutions.
In the future, we aim to extend our method to address dynamic real-world MOCOP instances, integrating domain-specific constraints, and improving generalizability in online environments.

\clearpage
\bibliographystyle{splncs04}
\bibliography{refs}
\end{document}